\documentclass[conference]{IEEEtran}
\usepackage{color}
\usepackage{cite}

\ifCLASSINFOpdf
   \usepackage[pdftex]{graphicx}
\else
\fi
\usepackage[cmex10]{amsmath}

\usepackage{stfloats}

\usepackage[lined]{algorithm2e}

\begin{document}
%
\title{Modeling Scalability of Distributed \\ Machine Learning}

\author{\IEEEauthorblockN{Alexander Ulanov}
\IEEEauthorblockA{Hewlett Packard Laboratories\\
Palo Alto, California, USA\\
Email: alexander.ulanov@hpe.com}
\and
\IEEEauthorblockN{Andrey Simanovsky}
\IEEEauthorblockA{Hewlett Packard Laboratories\\
Palo Alto, California, USA\\
Email: andrey.simanovsky@hpe.com}
\and
\IEEEauthorblockN{Manish Marwah}
\IEEEauthorblockA{Hewlett Packard Laboratories\\
Palo Alto, California, USA\\
Email: manish.marwah@hpe.com}
}


%


\maketitle

\begin{abstract}
Present day machine learning is computationally intensive and processes large amounts of data. 
It is implemented in a distributed fashion in order to address these scalability issues.
The work is parallelized across a number of computing nodes.
It is usually hard to estimate in advance how many nodes to use for a particular workload.
We propose a simple framework for estimating the scalability of distributed machine learning algorithms.
We measure the scalability by means of the speedup an algorithm achieves with more nodes.
We propose time complexity models for gradient descent and graphical model inference.
We validate our models with experiments on deep learning training and belief propagation.
This framework was used to study the scalability of machine learning algorithms in Apache Spark.
\end{abstract}


%
\IEEEpeerreviewmaketitle

\section{Introduction}
Nowadays a lot of machine learning workloads run in data centers.
Many of them, such as deep learning for speech recognition or computer vision, would likely take
weeks or months to run on a single node. 
Hence, they are typically run on a parallel and distributed platform. 
But how  many machines to use to deploy the workload?
In this paper we address that question.
There are two main scenarios where a practitioner encounters this question. 
(1) Given a workload, how many more machines are needed to decrease the run time by a certain amount?
(2) Given an increasing workload, how many more machines to add to keep the run time the same?
The first senario is referred to as {\em strong} scaling; the second one is called {\em weak} scaling.
The challenge is to come up with a good performance model of a machine learning method that accurately captures the processing time, while being easy to build and not requiring extensive instrumentation and data collection.

Performance models of distributed systems have been thoroughly studied in parallel algorithms community \cite{jaja1992introduction}.
In particular, they looked at speedup (how much faster a task runs on a multi-node system versus a single node one) and efficiency (what percentage of node resources are utilized in a multi-node system on average versus resources utilized in a single mode run).
Amdahl \cite{Amdahl:1967} suggested a law that describes how fast problems can be solved using parallel computations.
Gustafson \cite{Gustafson:1988} formulated consensus on what happens when larger problems are solved with more nodes.
Even though the parallel algorithms community often excluded communication overheads from their models, a computation-communication trade-off was well understood \cite{parhami2006introduction}.

Machine learning has become a widely accepted solution for many applications.
As big data becomes a reality, more and more of the algorithms are being parallelized. 
The most common way to address scalability is to implement the MapReduce paradigm.
Distributed machine learning frameworks (Spark ML, Tensorflow, and others) are there to help a practitioner.
Nonetheless, scaling of a machine learning task remains hit and trial.
That happens because scaling of machine learning algorithms requires parallelization of algorithms that were previously considered strictly sequential and practitioners do not know where to look for bottlenecks. 


We propose a simple framework for distributed computation models and validate it for diverse machine learning algorithms, which
include multiple neural network architectures, all of which require gradient-descent, and loopy belief propagation, a popular 
graphical model inference algorithm.
For these use cases we build theoretical models and validate them experimentally.
Our distributed machine learning modeling framework enables a practitioner to build a speedup plot for a machine learning 
algorithm and use it to estimate an optimal number of machines.
The framework views algorithms as combinations of computation and communication steps.
It uses algorithm-independent parallelization techniques.
The framework takes algorithm-specific computation and communication time complexity formulas as inputs, but unlike prior art, does not 
require any test runs or profiling of the algorithms.

The contributions of this work are the algorithm-independent framework, algorithm models for the use cases, 
and validation of the use cases.
The empirical results match well with the model estimates.
The framework was used to develop implementation of deep learning on Apache Spark and study scalability of it and other algorithms in Spark ML \cite{talk}.
Our framework (including validation results) is also available online in the form of iPython notebooks \cite{github}.

\section{Related work}


Most of the work on performance modeling and resource allocation for machine learning workloads focus on Hadoop/MapReduce.
In addition, they require profiling data and use complicated models.
In contrast, our method requires only hardware specification.

Gandhi et al.~\cite{gandhi2016autoscaling} define a fine-grained model for Hadoop/MapReduce.
The computation model is approximated by second-order polynomials and, thus, would not be accurate if the workload has a higher-order time complexity, such as deep learning.
Herodotou and Babu~\cite{herodotou2011} build models to answer what-if performance questions, however, they require collection of extensive profiling data.
The closest work to ours is Sparks et. al.~\cite{sparks2015} where they propose a cluster resource allocation estimator.
They have a similar model for parallel computation–computation time is inversely proportional to the number of nodes;
communication time is total communication multiplied by the number of nodes.
However, this accounts for only linear communication architectures and inaccurate for all-reduce, which is implemented in MPI, and other communication paradigms, such as
shuffle mechanism implemented in Hadoop MapReduce and Apache Spark.
Furthermore, their model requires experimental data to select appropriate parameters.
They take into account the framework overhead as sequential step (aka Amdahl fraction).
However, according to Scheiber~\cite{schreiber2014}, one could make it decline with increasing n, so that the sequential piece is irrelevant to scaling.
Another more recent work related to ours is Venkataraman et. al.~\cite{venkataraman2016ernest}.
Their model is similar to~\cite{sparks2015} with the addition of logarithmic dependency on the number of workers.
The model requires experimental data for parameter estimation.
Keuper and Pfreundt~\cite{keuper2016distributed} investigate the scalability of deep learning training and experimentally show that there is a communication bottleneck.
They also discuss different ways of organizing the communication and how the training can be scaled by increasing the amount of computations.
While their findings match with ours, they only focus on deep learning and don’t explicitly provide a modeling framework.

\section{Proposed Methodology}
\label{sec:PM}

Consider a distributed machine learning algorithm running on a cluster of homogeneous nodes. 
There are two components that we model: the machine learning algorithm and the framework for distributed computations. 
We assume the algorithm is implemented using the bulk synchronous parallel (BSP) framework \cite{valiant1990bridging},
comprising a series of \textit{supersteps}.
Each \textit{superstep} is a sequence of concurrent computation and communication steps with a synchronization 
barrier at the end.
Computation and communication are algorithm-dependent.
We define the distributed computation time complexity as
\begin{gather*}
t_{cp} = c(D) / n
\end{gather*}
where $n$ is the number of homogeneous ndoes; $c$ is the computation time complexity function, which depends on the input size $D$.
The communication time complexity is
\begin{gather*}
t_{cm} = f_{cm}(M, n)
\end{gather*}
where $M$ is the total number of messages sent through the communication medium.
$M$ depends on the algorithm design; the shape of $f_{cm}$ depends on the topology of the media.
The time complexity of a \textit{superstep} is determined as the sum of the two terms, since computation 
and communication steps do not overlap.
(We assume that the synchronization barrier is implicitly included in the computation.)
\begin{gather*}
t = t_{cp} + t_{cm}
\end{gather*}
We require a model, i.e. computation and communication time complexity formulas, for each algorithm.
That allows the scalability model to be accurate.

We use speedup to measure the effectiveness of a distributed machine learning algorithm:
\begin{gather*}
s(n) = t(1) / t(n)
\end{gather*}
where $t(1)$ is time complexity with one computing node, and $t(n)$ is time complexity with $n$ nodes.
We use speedup rather than the total time itself because, being a relative metric, speedup equation cancels out proportional systematic errors.
The algorithm is scalable if there exists $k$ such that $s(k)>1$.
The optimal number of nodes is $N = argmax s(n)$.

We can examine strong and weak scaling of a machine learning algorithm.
Strong scaling is when we fix the input size $D$ and vary the number of computing nodes.
Weak scaling is when we vary both the input size and the number of nodes.

\begin{figure}[h] 
  \includegraphics[width=0.5\textwidth]{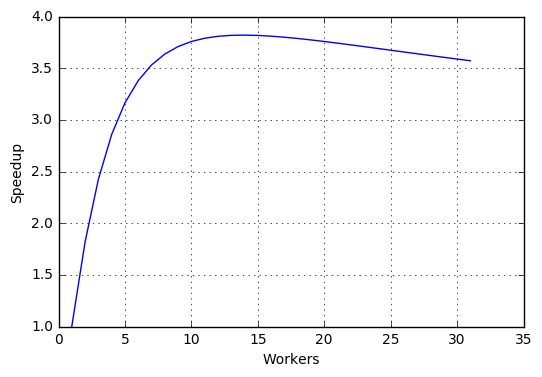}
  \caption{Example of the speedup }
  \label{fig:example}
\end{figure}

Consider an example. 
Suppose, we want to examine strong scaling of a machine learning algorithm.
Figure~\ref{fig:example} shows speedup depending on the number of nodes, or \textit{workers}.
Per-node computation time complexity decreases as the number of nodes grows.
At the same time, communication time complexity increases.
Consequently, speedup does not grow indefinitely and starts to decrease at around 14 nodes.
This is the point where the total execution time reaches its minimum.

Our simple approach does not require data gathering and profiling, but still provides useful and accurate results.
It allows practitioners to estimate optimal amounts of resources for efficient runs of machine learning workloads.
In the following sections we use speedup to analyze scalability of machine learning workloads.

\section{Use cases}
In this section we discuss specific machine learning workloads, gradient descent and graphical model inference.
Gradient descent is the most widely used optimization algorithm in machine learning.
It is used for training machine learning models for such tasks as click-through rate prediction in display advertising, image classification, and speech recognition.
On the other spectrum of machine learning methods are graphical models.
Many domains are naturally represented as graphs where nodes or edges hold some parameters that one wants to estimate.
The latter is accomplished with a particular probabilistic inference algorithm.
Graphical models are applied in such use-cases as computation of the importance of web-pages, product recommendations, 
and fraud detection.

The important property of the use cases is that they involve large training data.
Understanding the scalability of the use cases helps a practitioner to choose among a multitude of configuration options,
and save time and costs.

\subsection{Gradient descent}
Gradient descent is applied to a broad variety of machine learning models, including linear and non-linear models.
Widely used logistic regression and deep learning models are trained using gradient descent.
Gradient descent iteratively modifies parameters of the model in order to optimize a cost function, which is defined with respect to the parameters and the training set.
The gradient of the cost function is computed on each iteration. 
It depends on the current parameters of the model and the training set.
Then each parameter is summed with the gradient value multiplied by a small constant.
The iterations are repeated until the parameter values converge.
Batch gradient descent implies that gradient is computed using the whole training set on each iteration.
Stochastic gradient descent (SGD) uses a random sample on each iteration.
Mini-batch SGD uses a random mini-batch of examples.

Gradient descent and mini-batch SGD can be data parallel---each node computes the gradient in parallel using a part of the batch.
Then the results are collected to the master node.
The master node computes the update and broadcasts it to the workers.
Both communications can be organized as a tree in order to reduce their time complexity.

According to the proposed model, we can calculate the computation and communication time complexity for gradient descent as follows:
\begin{gather*}
t^{GD}_{cp}= (C \cdot S) / (F \cdot n)
\end{gather*}
where $n$ is the number of nodes with $F$ floating point operations per second (FLOPS); $C$ is the computation time complexity of gradient descent on one data point; $S$ the size of the batch;
\begin{gather*}
t^{GD}_{cm}=2 \cdot (32 \cdot W / B) \cdot log(n)
\end{gather*}
where $W$ is the number of 32-bit parameters in the machine learning model; $B$ is the connection bandwidth between computing devices; $2$ accounts for two-stage communication.
We do not account for the initialization time because the number of iterations until convergence is usually large.
Consequently, the initialization time can be ignored.

In Section~\ref{sec:TDLM} we discuss scalability of gradient descent and mini-batch SGD for deep learning.
\subsection{Graphical models}
\label{sec:GM}
Graphical models \cite{koller2009probabilistic} combine probabilistic dependency and reasoning with graph theory.
Vertexes in the graph represent random variables, while edges denote dependencies. 
The graph structure provides an elegant representation of the conditional independence relationships between variables.
An important task in these models is inferring the state of unknown or hidden variables. 
While exact inference is usually intractable for large models, approximate methods, such as Gibbs sampling or loopy belief propagation, are commonly used. 
In our analysis, we consider pairwise Markov random field (MRF) model, which is generic enough to represent any graphical model.
The inference methods are iterative and can be parallelized in different ways depending on the graph representation.
We build a model where vertexes are processed in parallel by multiple worker nodes.
The variable value for a vertex is collected remotely for all its neighbor vertexes that are processed by other workers.

According to the proposed model, we can define the computation time complexity for graph inference as follows:
\begin{gather*}
t^{GI}_{cp}= max_{i \in [1, n]}(E_i) \cdot c(S) / F
\end{gather*}
where $F$ is FLOPS; $max_{i \in [1, n]}(E_i)$ is maximum number of edges after distributing vertexes across $n$ workers; $c(S)$ is the number of computations of an algorithm given $S$ number of states in a variable.

The number of edges per worker can be estimated via Monte-Carlo-like simulation.
In order to do this, we randomly assign each vertex to a worker and add its degree to the total number of edges on the worker $E^{rnd}_i$.
In this way we count edges that connect vertexes from the same worker twice.
Taking into account that the number of vertexes on a worker is $V/n$; the maximum number of edges on a worker is $1/2\cdot(V/n-1)V/n$; the maximum number of edges in the whole graph is $V(V-1)/2$; and the probability of two vertexes to share an edge is $E / (V(V-1)/2)$, we estimate the number of the edges counted twice as: 
\begin{gather*}
E^{dup}= 1/2\cdot(V/n - 1)V/n\cdot E / (V(V-1)/2)
\end{gather*}
The estimated number of edges on a worker $i$ is $E_i=E^{rnd}_i - E^{dup}$.

The communication time complexity accounts for the variables that need to be collected remotely (replicated).
In case of the linear communication we have:
\begin{gather*}
t^{GI}_{cm}= 32 / B \cdot r \cdot V \cdot S
\end{gather*}
where $B$ is the connection bandwidth between workers; 32 is the number of bits per state; $r$ is the replication factor; $V$ is the total number of vertexes.

In section \ref{sec:PIGM} we discuss scalability of graphical model inference.

\section{Experiment results and validation}
We conducted a series of experiments to validate the proposed models for gradient descent and graphical model inference.
\subsection{Training of deep learning models}
\label{sec:TDLM}
Deep learning is a widely used machine learning method.
It is based on multi-layer artificial neural networks (ANN).
ANNs are statistical models that approximate functions of multiple inputs.
A network consists of interconnected ``neurons''.
A neuron produces an output by applying a function, optionally parametrized with weights, to its input.
For example, a neuron with a linear function returns a linear combination of the inputs with weights being the coefficients of the combination.
A neuron with a sigmoid function applies a logistic function to the input. 
Neurons are grouped into layers.
Typically, neurons of one layer can operate in parallel.
Neurons in different layers work sequentially.
For example, layers of sigmoid functions usually follow linear layers.
Networks with more than three layers are usually called ``deep''.

Deep learning model parameters are the weights that neurons use for transformations.
These parameters can be estimated with gradient descent on the cost function of the model prediction error on the training set.
The gradient is computed with the back propagation algorithm.
It involves three steps: forward propagation, back propagation of the cost function error, and computation of the gradient using the error.

The computation time complexity in terms of ``multiply-add'' operations for fully-connected layers can be estimated as $6 \cdot W$. 
(Each of the steps has two matrix multiplications per each network layer $2\cdot n_i\cdot m_i=2\cdot w_i$---where $n_i$ and $m_i$ are the number of column and rows in the weight matrix of layer $i$ and $w_i$ is the total number of weights in the layer).

The computational time complexity of a convolutional layer is $n \cdot (k \cdot k \cdot d \cdot c \cdot c)$---where $n$ is the number of feature maps; $k$ is the size of the map (assuming it is squared); $d$ is the depth of the input tensor; $c$ is the number of sliding windows for the feature map on the input.
$n$, $k$, and the depth $d_0$ of the input are given in the network configuration.
$d_i$ is calculated from the shape of the output of $(i-1)$-th layer.
$c=(l - k + b) / s + 1$---where $l$ is one side of the input (assuming it is squared); $k$ is its depth; $b$ is the border size; $s$ is stride; and $/$ is integer division.
$b$ and $s$ are defined by the network configuration.
The number of weights in the convolutional layer is $n \cdot (k \cdot k \cdot d + c \cdot c)$.
Bias (the number of weights is $c \cdot c$) is not commonly used for convolutional layers.

We used the above formulas to compute the number of weights and computations for deep learning models in our experiments.
We verified that it corresponds to the numbers listed in the papers describing these models.
Details are available online in our iPython notebooks~\cite{github}.

Table~\ref{tab:networks} contains the properties of the neural networks for our experiments.
The fully connected network that we use is one of the most accurate networks used for MNIST handwritten character recognition~\cite{dan2010deep}.
It has five hidden layers (2500, 2000, 1500, 1000, and 500 neurons), 784 inputs, and 10 outputs.
We also consider Inception v.3, a convolutional network for ImageNet image classification challenge~\cite{szegedy2015rethinking}.
The table also lists the number of ``multiply-add'' computations required for the forward passes for these networks.

\begin{table}[h]
  \centering
  \caption{Network configurations}
  \label{tab:networks}
  \begin{tabular}{ | c | c | c | c | }
    \hline
    Network (Task) & Parameters & Computations \\ \hline
    Fully connected (MNIST) & $12\cdot 10^6$ & $24\cdot 10^6$  \\ \hline
    Inception v.3 (ImageNet) & $25\cdot 10^6$ & $5\cdot 10^9$ \\
    \hline
  \end{tabular}
\end{table}

\begin{figure}[h] 
  \includegraphics[width=0.5\textwidth]{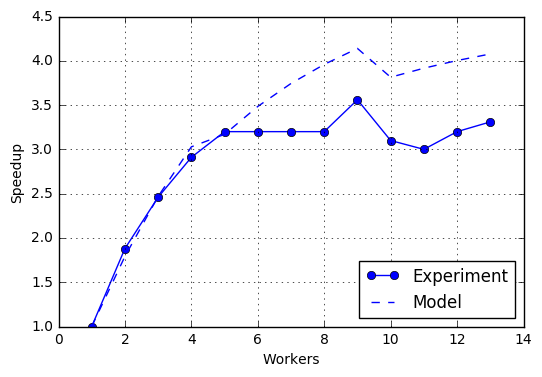}
  \caption{Speedup of one iteration for fully connected ANN training}
  \label{fig:ann}
\end{figure}

Figure~\ref{fig:ann} shows the experimental and theoretical speedup of fully-connected ANN training.
Markers represent experiments.
We used the Apache Spark~\cite{spark} 64-bit implementation of ANN in our experiments.
Spark was running on a cluster.
Each Spark worker was run on a dedicated node.
Master also had a dedicated node.
Each node is Xeon E3-1240 with 16GB of RAM and 1~Gbit/s network under RedHat Linux 6.3.
The CPU has 211.2~GFLOPS~\cite{intel}.
We assume that one can reach at most 80\% of the peak FLOPS.

According to the proposed model of gradient descent and the above estimations of computations, the computation time complexity for a fully connected ANN is:
\begin{gather*}
t_{cp}= (6 \cdot W \cdot S) / (F \cdot n)
\end{gather*}
where $n$ is the number of equal computing nodes with $F=0.8\cdot 105.6\cdot 10^9$ double precision FLOPS; $W=12\cdot 10^6$ is the number of 64-bit parameters; $S$ is the size of the batch.
Spark uses batch gradient descent, that is, the batch size equals to the size of the dataset, which is 60000 samples.
Two stage communication for gradient distribution and aggregation happens with two different protocols in Spark.
Distribution of parameters is implemented with a torrent-like protocol.
Aggregation is done in two waves.
First wave is done for the square root number of the nodes and the second wave is done among the others.
The communication time can be defined as:
\begin{gather*}
t_{cm}= (64 \cdot W / B) \cdot log(n) + 2 \cdot (64 \cdot W / B) \cdot \lceil \sqrt{n} \rceil
\end{gather*}
where bandwidth is $B=10^9$~bit/s.

The model provides very good estimation of experimental results up to five workers.
It also has the same peak, which accounts for the way the gradient communication is implemented in Spark.
The model suggests that the optimal number of workers is nine.
Adding more workers does not provide any speedup due to communication overhead.
The mean absolute percentage error (MAPE) of the model estimates is only 13.7\%.

\begin{figure}[h] 
  \includegraphics[width=0.5\textwidth]{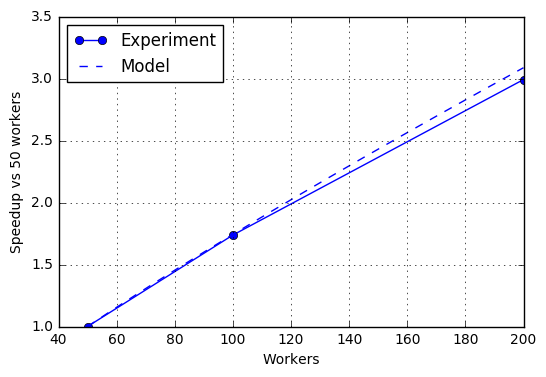}
  \caption{Speedup of processing time per training instance for convolutional ANN training (relative to 50 nodes)}
  \label{fig:cnn}
\end{figure}

Figure~\ref{fig:cnn} shows experimental and theoretical speedup of convolutional ANN training.
We used the experimental results from the paper by Chen et al. ~\cite{chen2016revisiting} for speedup vs 50 workers.
Chen and others conducted experiments with a GPU cluster of nVidia K40 GPU and TensorFlow software~\cite{abadi2016tensorflow}.
They list the number of parameters $W=25\cdot 10^6$ and the computation time complexity $C=3\cdot 5\cdot 10^9$.
Each worker has nVidia K40 GPU with 4.28~TFLOPS.
We assume that it can reach at most 50\% of peak performance.
We compared our model predictions with their experiments with synchronous minibatch SGD.

The authors of \cite{chen2016revisiting} explored the scalability of synchronous and asynchronous mini-batch SGD in terms of model convergence.
Each node computes a gradient for a fixed batch size.
Adding more nodes increases the size of the effective batch and the training scales in terms of instances per second processed.
This is an example of weak scaling.
We calculate the speedup of processing of one instance to account for that.

The time complexity of processing of one instance is:
\begin{gather*}
t= ((C \cdot S) / F + 2 \cdot (32 \cdot W / B) \cdot log(n)) / n
\end{gather*}
where mini-batch size $S=128$, which is a typical choice for a single worker, and $B=10^9$~bit/s.
We assume that gradient aggregation uses logarithmic model of communication.
Such assumption allows infinite weak scaling, i.e. adding more workers always increases single instance speedup.
The linear communication model allows only finite scaling: after enough workers added, the speedup remains constant.
Linear communication model only scales when the communication time for one worker is less than the computation time for it.

According to Figure~\ref{fig:cnn}, the proposed model provides very close estimation of empirical speedup.
MAPE is 1.2\%.
It can help to plan experiments for exploring the convergence of deep learning training.

\subsection{Belief propagation}
\label{sec:PIGM}

Belief propagation (BP)~\cite{pearl1988probabilistic} is a popular message-passing based algorithm for computing marginals of random variables in a graphical model, such as a Markov random field.
It provides exact inference for trees, and approximate inference for graphs with cycles (in which case it is referred to as loopy belief propagation).
Even though loopy belief propagation is an approximate algorithm with no convergence guarantees, it works well in practice for many applications~\cite{murphy1999loopy} such as coding theory, image denoising, malware detection.
  

The loopy belief propagation algorithm works in two steps: (i) based on the messages from its neighbors, a vertex updates its own belief; and (ii) based on its updated belief, a vertex sends out messages to its neighbors.
In the \textit{update step} (i), a vertex belief is a product of its prior belief and messages about its variable received from its neighbors.
In the \textit{send step} (ii), a vertex generates a new message for each of its neighbors about the neighbor's variable. In the message, the vertex marginalizes over the values of the variable received except for the value received from the neighbor.
These steps are repeated until convergence.

According to the proposed model, the computation time complexity for BP is:
\begin{gather*}
t_{cp}= max_{i \in [1, n]}(E_i) / (F \cdot n) \cdot (S + 2\cdot(S + S^2))
\end{gather*}
where $n$ is the number of equal computing devices with $F$~FLOPS; $S=2$. 
$max_{i \in [1, n]}(E_i)$ is estimated as discussed in Section~\ref{sec:GM}.

\begin{figure}[h] 
  \includegraphics[width=0.5\textwidth]{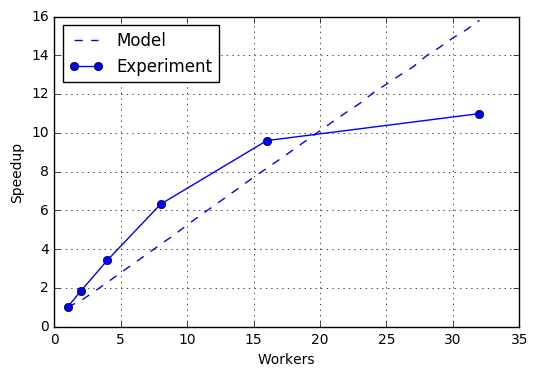}
  \caption{Speedup of the BP algorithm with a graph of 16M vertexes}
  \label{fig:graph10m}
\end{figure}

Figure~\ref{fig:graph10m} shows the results of the experiments with a graph containing 16,259,408 vertexes, 99,854,596 edges, and a maximum degree of 309,368.
The graph is based on real DNS data traffic in a large enterprise.
Theoretical curves correspond to the speedup with $t_{cp}$ estimated with a Monte-Carlo-like simulation.
Experimental curves were obtained with our implementation of BP in GraphLab~\cite{graphlab}, which ran on an HP ProLiant DL980 server with 80 CPU cores at 1.9 GHz and 2 TB of memory. 
We made a simplifying assumption that communication time complexity is negligible because all communications happen in the shared memory.
$F$ is factored out in the speedup formula due to this assumption.
MAPE is 25.4\%.
As one can see on the charts, random vertex assignment to workers turns out to be a conservative estimate for configurations with few workers.
However, execution overhead takes over with larger number of workers.
We conducted experiments for smaller graphs with 1.6M, 165K, and 16K vertexes~\cite{github}. MAPE is 26\%, 19.6\%, and 23.5\% correspondingly.

\section{Conclusion and Future Work}

We presented a methodology for building performance models of distributed machine learning algorithms,
which we applied to two machine learning use cases, gradient descent and graphical model inference.
We validated the framework with experiments on neural networks and belief propagation. 
The results showed a close match between our models and the emperical data.  

There are several opportunities for future research.
One direction is enriching the machine learning computation models.
In that respect, we consider building a model for asynchronous algorithms, such as asynchronous gradient descent \cite{NIPS2011_0485}.
We also are interested in looking into parallelization-convergence trade-offs specific to machine learning.
For example, gradient descent parallelization techniques pay for parallelism with algorithmically slower convergence or convergence to a worse local optimum.
Finally, incorporating a feedback loop from experiments would also be useful as we found out with modelling belief propagation calculations.

Another future work direction is popularizing the adoption of such models in the industry.
We strongly believe that the kind of simple, almost back-of-the-envelope scalability estimations that are presented in the paper should precede distributed implementations (and may sometimes prevent them!).
The possible solution is to integrate the estimation software with such tools as Spark, Hadoop, and Tensorflow.

\section*{Acknowledgment}

The authors would like to thank Rob Schreiber from Hewlett Packard Labs and Xiangrui Meng from Databricks for inspiring discussions.
We would also like to thank Carlos Zubieta for help with the belief propagation experiments.




%
%
%

\end{document}